%% file: main.tex
\documentclass[10pt,journal,compsoc]{IEEEtran}
\ifCLASSOPTIONcompsoc
  \usepackage[nocompress]{cite}
\else
  \usepackage{cite}
\fi
\ifCLASSINFOpdf
\else
\fi
\usepackage{times}
\usepackage{epsfig}
\usepackage{graphicx}
\usepackage{amsmath}
\usepackage{amssymb}
\usepackage{graphicx}
\usepackage{amsmath}
\usepackage{amssymb}
\usepackage{booktabs}
\usepackage{tabu}

\usepackage[pagebackref]{hyperref}
\usepackage{xcolor}
\usepackage{array}
\usepackage{epsfig}
\usepackage{epstopdf}
\usepackage{booktabs}
\usepackage{algorithm}
\usepackage{algpseudocode}
\usepackage{lipsum}
\usepackage{multirow}
\usepackage{textcomp}
\usepackage{gensymb}
\usepackage{pifont}
\begin{document}
%
\title{GaitSTR: Gait Recognition with Sequential Two-stream Refinement}
%
%
%
%

\author{Wanrong~Zheng*, 
        Haidong~Zhu*, 
        Zhaoheng~Zheng,
        and~Ram~Nevatia,~\IEEEmembership{Fellow,~IEEE}
\IEEEcompsocitemizethanks{\IEEEcompsocthanksitem The first two authors contributed equally. %
\IEEEcompsocthanksitem W. Zheng, H. Zhu, Z. Zheng and R. Nevatia are with the Department
of Computer Science, University of Southern California, CA, 90089.\protect\\%
E-mail: \{wanrongz,haidongz,zhaoheng.zheng,nevatia\}@usc.edu.}
\thanks{Manuscript received November 14, 2023.}
}

%
%

\markboth{IEEE Transactions on Biometrics, Behavior, and Identity Science,~Vol.~14, No.~8, August~2015}%
{Zheng \MakeLowercase{\textit{et al.}}: GaitSTR: Gait Recognition with Sequential Two-stream Refinement}

\IEEEtitleabstractindextext{%
\input{sections/0-abs}

\begin{IEEEkeywords}
Gait Recognition, Skeletons Representation, Joint and Bone, Temporal Refinement.
\end{IEEEkeywords}}

\maketitle

\IEEEdisplaynontitleabstractindextext

%
\IEEEpeerreviewmaketitle

\IEEEraisesectionheading{\section{Introduction}\label{sec:introduction}}

%
%
%
%

\input{sections/1-intro}
\input{sections/2-related}

\input{sections/3-method}

\input{sections/4-exp}

\input{sections/5-conclusion}


%



\ifCLASSOPTIONcompsoc
  \section*{Acknowledgments}
\else
  \section*{Acknowledgment}
\fi
\input{sections/acknowledgement}

\ifCLASSOPTIONcaptionsoff
  \newpage
\fi



%

\bibliographystyle{IEEEtran}
\bibliography{IEEEabrv, egbib}

%

\begin{IEEEbiography}
[{\includegraphics[width=1in,height=1.25in,clip,keepaspectratio]{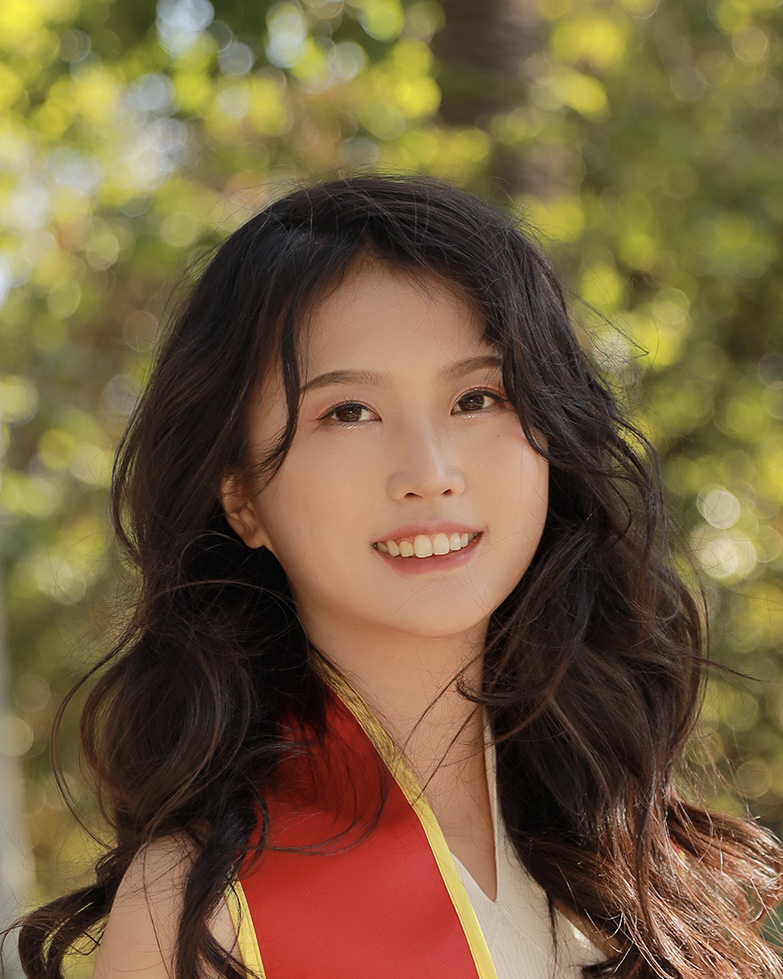}}]{Wanrong Zheng} received her MS degree in Computer Science from the University of Southern California (USC) in 2023. Before her time at USC, she worked at SenseTime as a full-time research engineer from 2019 to 2021. Her research interests are computer vision and deep learning, focusing on multi-modal perception and explainable artificial intelligence.
\end{IEEEbiography}

\begin{IEEEbiography}[{\includegraphics[width=1in,height=1.25in,clip,keepaspectratio]{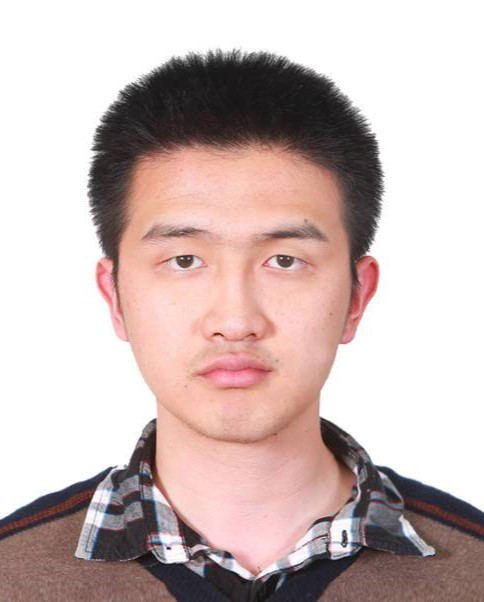}}]{Haidong Zhu} received the BS degree in Electronic Information Science and Technology from Tsinghua University in 2019. He is currently pursuing the Ph.D. degree with the Department of Computer Science, University of Southern California. His interests include computer vision, focus on 3-D vision and its application for downstream vision tasks.

\end{IEEEbiography}

\begin{IEEEbiography}[{\includegraphics[width=1in,height=1.25in,clip,keepaspectratio]{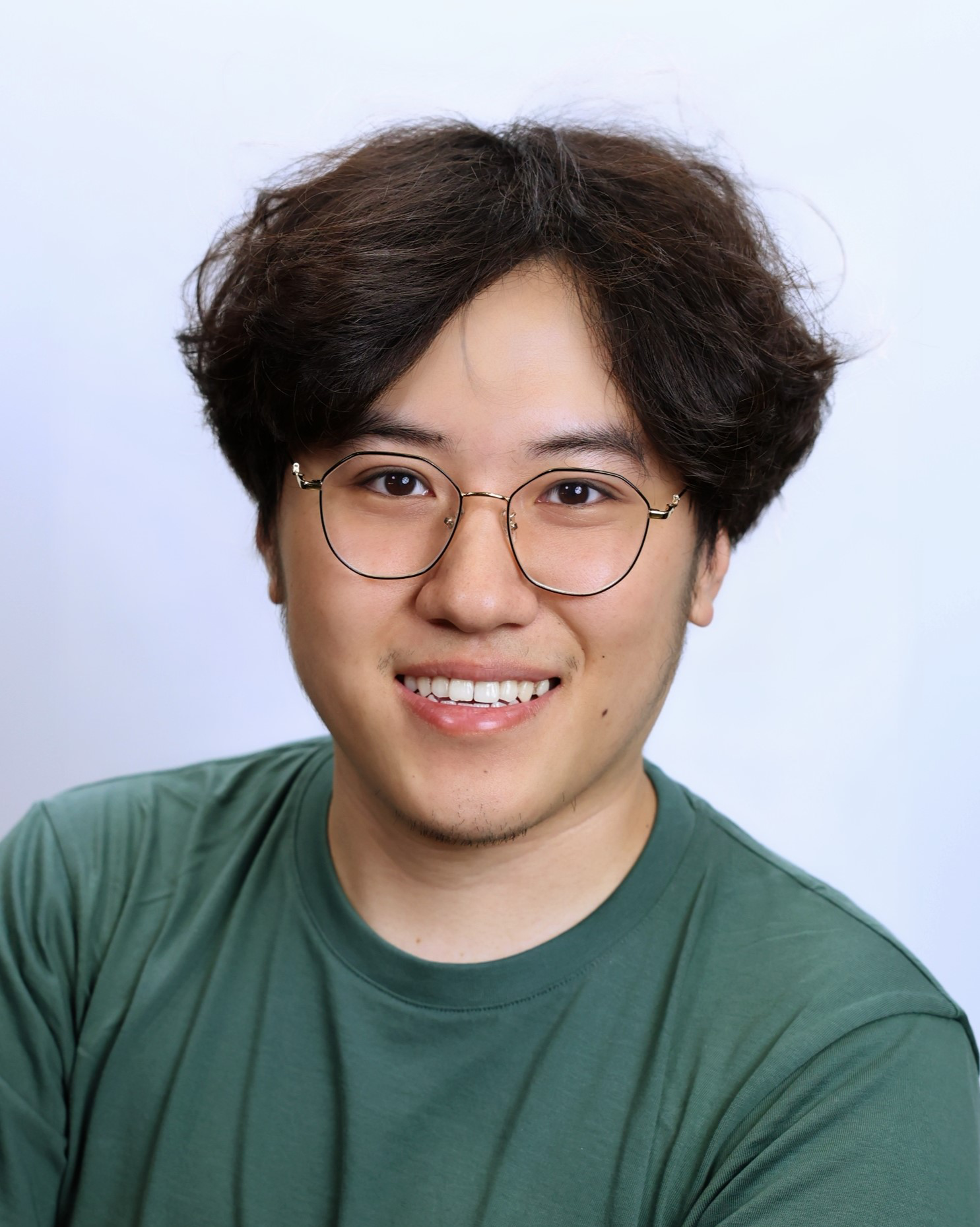}}]{Zhaoheng Zheng} is currently pursuing his Ph.D. in the Computer Science Department, University of Southern California. His research interest includes joint understanding of vision-langauge corpora and large multimodal models. Prior to USC, he obtained his M.S. in Computer Science from University of Michigan, Ann Arbor and his B.Eng. in Computer Science and Technology from Tsinghua University. 
\end{IEEEbiography}


\begin{IEEEbiography}[{\includegraphics[width=1in,height=1.25in,clip,keepaspectratio]{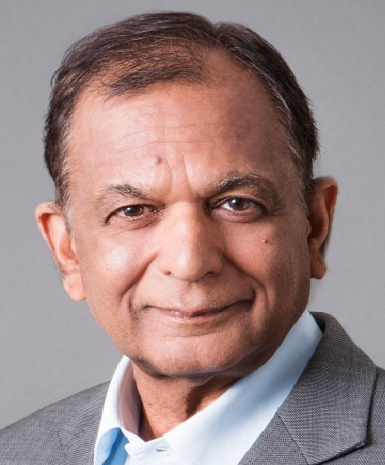}}]{Ram Nevatia} (Fellow, IEEE) received his PhD degree in Electrical Engineering from Stanford University, California. He has been with the University of Southern California since 1975, where he is currently Fletcher Jones Professor of computer science and Professor of electrical and computer engineering. He has made contributions to various areas of computer vision including object recognition, image grounding and action recognition.
\end{IEEEbiography}




\end{document}

%% file: sections/0-abs.tex
\begin{abstract}
Gait recognition aims to identify a person based on their walking sequences, serving as a useful biometric modality as it can be observed from long distances without requiring cooperation from the subject. In representing a person's walking sequence, silhouettes and skeletons are the two primary modalities used. Silhouette sequences lack detailed part information when overlapping occurs between different body segments and are affected by carried objects and clothing. Skeletons, comprising joints and bones connecting the joints, provide more accurate part information for different segments; however, they are sensitive to occlusions and low-quality images, causing inconsistencies in frame-wise results within a sequence. In this paper, we explore the use of a two-stream representation of skeletons for gait recognition, alongside silhouettes. By fusing the combined data of silhouettes and skeletons, we refine the two-stream skeletons, joints, and bones through self-correction in graph convolution, along with cross-modal correction with temporal consistency from silhouettes. We demonstrate that with refined skeletons, the performance of the gait recognition model can achieve further improvement on public gait recognition datasets compared with state-of-the-art methods without extra annotations. 
\end{abstract}

%% file: sections/1-intro.tex
\begin{figure}[t]
    \centering
    \includegraphics[width=\linewidth]{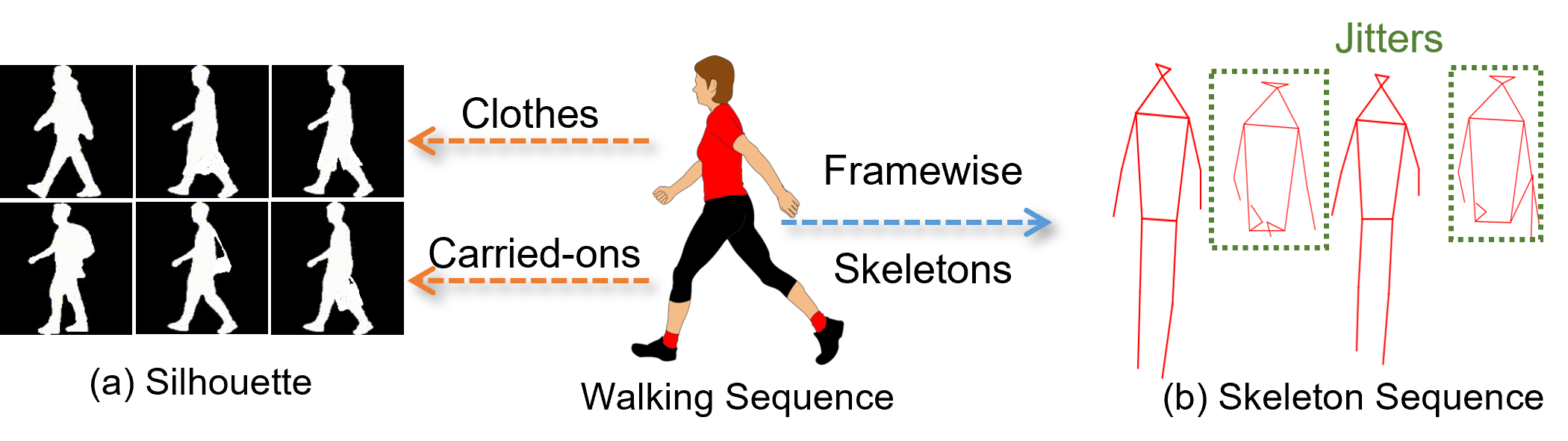}
    \caption{Visualization of the (a) silhouette and (b) skeleton sequence used for gait recognition. Silhouettes show different contours with different clothes and carried-on objects, while the skeletons suffer from jittery detection results in the video.}
    \label{fig:example}
\end{figure}

\IEEEPARstart{G}{ait} recognition~\cite{he2018multi,song2019gaitnet,wu2016comprehensive,yu2017invariant} is to identify the person present in a walking sequence. Different from other modalities, gait has the advantage of being able to be observed from a long distance and without the subject's cooperation. For gait recognition, researchers have developed silhouette-based methods, such as GaitSet~\cite{chao2019gaitset}, GaitPart~\cite{fan2020gaitpart}, GaitGL~\cite{lin2021gaitgl}, \textit{etc.}, and skeleton-based methods like GaitGraph~\cite{teepe2021gaitgraph}. However, both input modalities exhibit certain deficiencies. Binarized silhouettes suffer from variations due to clothing and carried objects, as shown in Figure~\ref{fig:example} (a), introducing external ambiguity, with segmented parts of a binarized silhouette being unavailable. Skeletons, on the other hand, include inconsistencies across frames in a sequence due to erroneous joint predictions, as depicted in Figure~\ref{fig:example} (b), thereby reducing the accuracy of gait recognition.%

In this paper, we propose the fusion of silhouette sequences with skeletons, harnessing the advantages of both modalities by refining the skeletons using silhouette sequences. Given that jitters in the detected skeletons are confined to a few frames isolated from the entire sequence, they lack temporal consistency with their neighboring frames~\cite{zeng2021smoothnet}. Simple temporal smoothing, however, can introduce further confusion for gait recognition as the generated skeletons create new poses inconsistent with the current sequence. On the other hand, silhouettes for neighboring frames exhibit better temporal consistency due to minor changes in adjacent image conditions. We enhance the quality of skeletons by employing silhouettes to rectify the jitters while retaining necessary identity information for more accurate gait recognition.

We introduce \textit{GaitSTR}, a \textbf{S}equential \textbf{T}wo-stream \textbf{R}efinement method, to refine the skeletons and combine them with silhouettes for gait recognition based on GaitMix and GaitRef~\cite{zhu2023gaitref}. When extracting the silhouette and skeletons from the same walking sequence, the temporal consistency between the two modalities is capable of providing guidance for each other: when silhouettes are not preserving useful boundaries of the images, skeletons can furnish the positions of the joints for pose estimation and recognition, and when the detection results of skeletons are unreliable, silhouettes can provide the pose information for the current frame in the sequence.

To refine the skeletons, we introduce  two-level fusion: an internal fusion within skeletons and a cross-modal correction with the temporal guidance from the silhouettes describing the same walking sequences. As skeletons can be decoupled into two different representations~\cite{shi2019two,zhu2022temporal}, joints, and bones, we incorporate self-correction between the frame-wise joints and bones for increased consistency. Introduction of the bones, in addition to joints, is to provide more connectivity as the GCN~\cite{yan2018spatial} primarily focuses on the position of the joints and does not explicitly explore the distances between the nodes other than binarized connectivity. To refine these two representations, we incorporate two different spatial-temporal graph convolution branches~\cite{yan2018spatial} for these two modalities, with the same number of layers and dimensions at each level. After each graph convolution operation, we utilize a self-correction residual block to forward the information of the joints and bones, and add the information to the other branch residually~\cite{he2016deep} between the same level of layers. 

In addition to the internal fusion of skeletons, we further introduce the cross-modal fusion between the silhouette and skeletons, combining the encoded silhouette features with the encoded frame-wise skeleton features to predict the relative changes for joints and bones in the skeletons. Since the gait pattern should be consistent for the same person, features from the silhouettes and skeletons describing the same walking sequence should also be consistent, facilitating the refinement of the skeletons with encoded silhouette features. Moreover, the sequence-level silhouette feature aids the frame-level skeletons for each frame in understanding its corresponding poses without losing identity information, as the temporal feature for the person is consistent and shared across all the frames in the same walking sequence.

With the predicted changes to the points, we reintegrate them into the original skeleton sequence and employ the skeleton encoder to extract the skeleton feature. We then concatenate this feature with the silhouette feature to predict the identity of the sequence using the refined skeletons. We compare \textit{GaitSTR} with baseline multimodal gait recognition methods that use skeletons and silhouettes, including GaitMix~\cite{zhu2023gaitref}, which concatenates the silhouette and skeleton features, and GaitRef~\cite{zhu2023gaitref}, which uses silhouette features to refine joints. We show that skeleton refinement across the skeleton and silhouettes aids the final gait recognition, while adding internal correction within skeletons yields the best performance. We assess our method on four public datasets: CASIA-B~\cite{yu2006framework}, OU-MVLP~\cite{takemura2018multi}, Gait3D~\cite{zheng2022gait}, and GREW~\cite{zhu2021gait}. Our findings demonstrate that the refined skeletons, when combined with silhouettes, outperform other state-of-the-art gait recognition methods that utilize skeletons and silhouettes.

In summary, our contributions are: 1) we introduce \textit{GaitSTR} which combines skeletons and silhouettes in an end-to-end training framework for gait recognition networks, 2) we incorporate the two different representations, joints and bones, for enhanced skeleton correction through self-fusion within skeletons, and 3) we use the consistency between silhouettes and skeletons to assist in correcting jitters in skeletons without additional supervision.

This paper is an extension of a conference version paper~\cite{zhu2023gaitref}. The novel contributions of this work are as follows.
\begin{enumerate}
    \item In addition to using joints as the skeleton representations in {GaitMix} and {GaitRef}~\cite{zhu2023gaitref}, we jointly utilize joints and bones as the representations. Unlike the single modal, the joints and bones represent different attributes of the skeletons, complementing each other.
    \item Besides the cross-modal fusion between silhouette and skeletons for refinement, we also introduce fusion between joints and skeletons. We demonstrate that feature integration and refinement provide a more comprehensive understanding between each level of features, and yield more consistent feedback to the skeleton representation, which results in improved gait recognition accuracy.
    \item We include additional experiments show that \textit{GaitSTR}, an extension of {GaitRef} for the contribution of different modalities in skeletons and silhouettes.
\end{enumerate}

%% file: sections/2-related.tex
\section{Related Work}

\textbf{Gait Recognition} aims to ascertain the corresponding identity of a person from their walking pattern. Given privacy concerns associated with RGB images, gaits are typically captured as two representations, silhouettes~\cite{yu2006framework,takemura2018multi,dou2023gaitgci,wang2023dygait,ma2023dynamic,huang20213d,fan2023opengait} and skeletons~\cite{an2020performance,sun2023trigait,zhu2023gaitref,Guo_2023_ICCV,huang2023condition,cui2023multi}. Silhouettes document the boundary map of human segmentation. To mitigate the impact of appearance variants on human shapes, researchers have focused on part-based and body-shape reconstruction methods for gait recognition. GaitSet~\cite{chao2019gaitset} and GLN~\cite{hou2020gln} introduce set pooling and extract set features in the sequence. GaitPart~\cite{fan2020gaitpart} and GaitGL~\cite{lin2021gaitgl} partition the image into various small patches, utilizing local features to reduce the impact of appearance variants. Beyond directly mining identity information from silhouettes, ModelGait~\cite{li2020end}, Gait3D~\cite{zheng2022gait}, PSE~\cite{zhu2023sharc} and Gait-HBS~\cite{zhu2023gait} focus on 3-D shape reconstruction to assist identification from sequences. DyGait~\cite{wang2023dygait} and GaitGCI~\cite{dou2023gaitgci} explore temporal information between different frames to better model and understand dynamic gait patterns.

In addition to mining identity from silhouette sequences, some researchers~\cite{liao2020model,teepe2021gaitgraph} have focused on using skeletons instead of silhouettes for gait recognition. Compared to the body contours represented by silhouettes, skeletons only encompass the joints and can eliminate the impact of body shapes as well as the appearance of the person. GaitGraph~\cite{teepe2021gaitgraph} employs the HRNet~\cite{wang2020deep} for joint detections and utilizes the generated pose sequence for recognition. PoseGait~\cite{liao2020model} segregates the gait sequence into pose, limb, angle, and motion, subsequently analyzing the movements for each skeleton based on these four features independently before amalgamating them for gait recognition. In combining silhouettes and skeletons, Wang \textit{et al.}~\cite{wang2022two} directly concatenate the two features, which still endures erroneous joint detections. Some existing works, such as GaitRef~\cite{zhu2023gaitref} and GaitMixer~\cite{pinyoanuntapong2023gaitmixer}, discuss the combination and integration of silhouettes with skeletons directly. In our work, we explore different levels of temporal fusion within skeletons and across various modalities.

\textbf{Pose Estimation and Refinement} focus on extracting the human body poses and refinement. With the development of transformers, pose estimation is also transforming from CNN-based networks~\cite{xiao2018simple,cao2019openpose} to transformer backbone networks~\cite{Li_2021_CVPR,li2021tokenpose,yang2021transpose,xu2022vitpose}.
Pose estimation has experienced rapid development from CNNs~\cite{xiao2018simple} to vision transformer networks. Early works treat the transformer as a better decoder~\cite{Li_2021_CVPR,li2021tokenpose,yang2021transpose,YuanFHLZCW21}. Although the frame-level pose estimation accuracy is becoming more and more accurate, directly applying these methods to tasks with solid temporal relations, such as gait recognition, may introduce extra uncertainty with inaccurate joint predictions. For the sequence with strong temporal patterns, HuMoR~\cite{rempe2021humor} corrects the joint prediction of the person with the previous pose, and SmoothNet~\cite{zeng2021smoothnet} filters the jitters in the whole sequence with analysis for the first and second deviation of the position for each point. These methods can fix some slight jitters and noise in the long sequence but still suffer when the poses for a long sequence are inaccurate. For the task of gait recognition with temporal repeated patterns, even with inaccurate predictions for the long sequence, the model should still fix the joints with the consistent moving pattern of the same person, which these existing methods cannot achieve.

%% file: sections/3-method.tex
\section{Method}

\begin{figure*}[t]
    \centering
    \includegraphics[width=\linewidth]{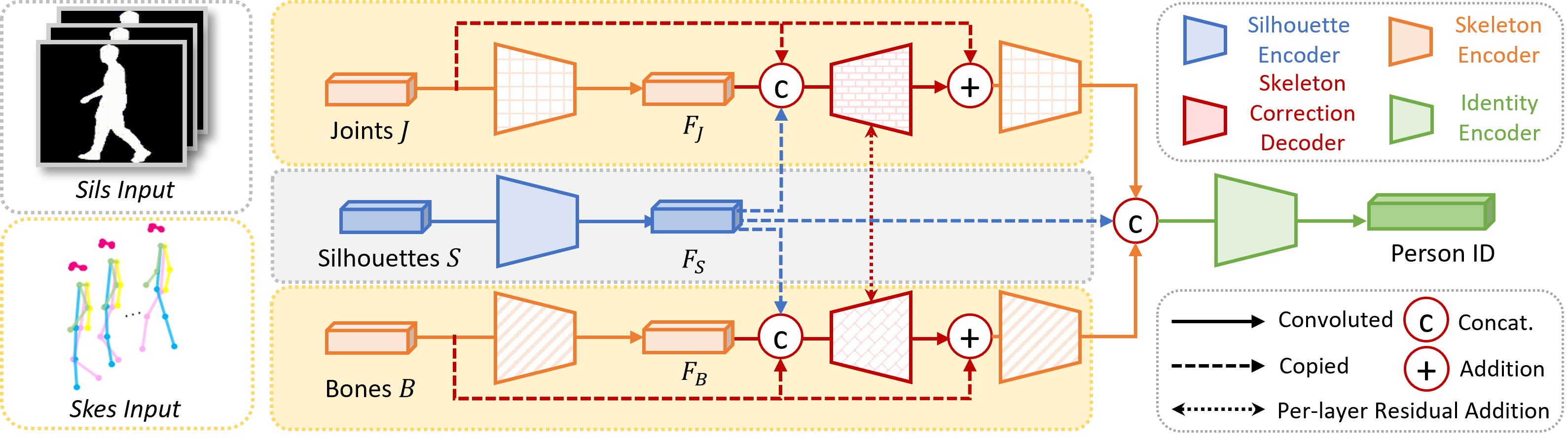}
    \caption{Our proposed architecture for GaitSTR. Trapezoids consists of trainable modules, and modules of the same color and fill-in patterns in the same model share the weights. Dashed lines represent the operation of feature copying. $S$, $J$, and $B$ are the input silhouettes, joints, and bones, respectively. $F_S$ represents silhouette features, while $F_J$ and $F_{B}$ represent joint and bone features for skeleton representations. }
    \label{fig:pipeline}
\end{figure*}

To recognize the person's identity, we combine silhouettes and skeletons for recognition. For silhouettes, we use the binarized body boundary images as input, and for skeletons, we take both bones and joints into consideration. Motivated by the bones used in skeleton-based action recognition~\cite{zhu2022temporal,xu2023language}, the introduction of bones emphasizes on the connections between the joints, while the joint-based graph convolution focuses largely on the nodes instead of the connection between them.

Given silhouettes $S$ along with the joints $J$ and bones $B$ of the skeleton for the person $p$, gait recognition is to match the identity with the people in a pool $P=\{p_n\}_{n=1,2,...}$, where $n$ is the candidate identity. We encode $S$, $J$ and $B$ to their corresponding embeddings and concatenate them together to find the nearest sample in $P$ in the embedding space. In this section, we first discuss the baseline combination of the bones in addition to the joints and silhouettes in Section~\ref{sec:gaitmix}. We then present our proposed method \textit{GaitSTR} to refine the input skeleton for gait recognition in Section~\ref{sec:gaitref}, along with objectives and details for training in Section~\ref{sec:obj}. We show our proposed architecture in Figure~\ref{fig:pipeline}.

\subsection{Baseline: Multimodal Gait Recognition}\label{sec:gaitmix}
 
We build the multimodal gait recognition model as our baseline to combine information from different input modalities, including skeleton, joint and silhouettes. We employ two encoders: a silhouette feature encoder designed for encoding the silhouette $S$, and a skeleton feature encoder tasked with projecting the dual representations of the input raw skeletons, namely joints $J$ and bones $B$, into their corresponding embedding spaces. Features generated from these two encoders are concatenated together and used for gait recognition.

\textbf{Silhouette Feature Encoder.} To extract the identity features from input sequential silhouette sequences, we use a silhouette feature encoder to convert the input silhouette sequence $S$ to the corresponding output identity feature $F_S$. We have three steps for the silhouette feature encoder: convolution feature extraction, temporal pooling, and horizontal pooling. With the binary silhouette input sequence $S = \{s_i\}_{i=1,..,t}$, where $i$ is the temporal stamp and $t$ is the overall frame number, we apply a convolution network to extract the framewise feature $f_{i}$ at frame $i$. $f_i$ is an $M$-by-$ N $-by-$ C$ matrix, where $M$ and $N$ are the height and width of the convoluted output features, and $C$ is the channel number from the output of the last convolution layer.

With the framewise feature $f_i$, we use a max pooling layer for the temporal fusion and combine the feature into a single $M$-by-$ N $-by-$ C$ output as temporal pooling. Since $f_i$ still includes the spatial features for each segment, we follow~\cite{fu2019horizontal} and apply horizontal pyramid pooling with scale $P$ as 5. The output of the feature is a $2^{P-1} $-by-$ C$ feature vector after horizontal pooling. The architecture of each component can be found in the implementation details in Section~\ref{sec:exp}.

\textbf{Skeleton Feature Encoder.} In addition to the silhouette encoder, we deploy a skeleton feature encoder in parallel. This encoder processes the input skeleton sequences, consisting of joints $J=\{j_i\}_{i=1,...,t}$ that record the position of each keypoint of the body, and bones $B=\{b_i\}_{i=1,...,t}$ that are represented as vectors denoting the numerical directional relationship between two connecting joints, into their associated embeddings denoted as $F_J$ and $F_B$. To represent both the joint and bone branches, we utilize two identical graph convolution networks; however, these networks do not share weights.

With the $N$-by-$K$-by-$2$ matrix to depict the 2-D skeletons of each frame, where $K$ (either $K_J$ for joints or $K_B$ for bones) represents the number of points or connections of the skeletons, we implement a multi-layer spatial-temporal graph convolution network~\cite{yan2018spatial} for graphical feature extraction for each of them. By transforming the input from dimensions $N$-by-$K$-by-$2$ to $N$-by-$K$-by-$C$ as the pre-frame per-node feature matrix, we then conduct average pooling over both the temporal and node dimensions and produce two final $1$-by-$C$-length vectors, $F_J$ and $F_B$, representing the features of the sequential skeleton pertaining to the input joints and bones. We concatenate these two vectors along their first dimension and generate a $2$-by-$C$ feature as the output of the skeleton feature encoder, along with the $2^{P-1}$-by-$C$ silhouette feature to represent the input sequence for recognition.

\subsection{GaitSTR: Sequential Two-stream Refinement}\label{sec:gaitref}
In addition to fusing features from skeletons and silhouettes for gait recognition, \textit{GaitSTR} introduces encoded features from silhouettes to improve the temporal consistency of skeletons, thereby enhancing the quality of skeletal data. The consistency across the two representations addresses framewise jitters in skeleton generation, ensuring a smoother and more accurate skeletal representation. Conversely, refined skeletons contribute to the silhouette analysis by minimizing the impact of appearance variants on gait recognition. Besides the two primary encoders employed for multimodal gait recognition, \textit{GaitSTR} incorporates two additional modules: a skeleton correction network, which rectifies framewise errors in skeleton predictions for joints and bones, and a cross-modal adapter, which bridges the gap between joint and bone representations to bolster the robustness of gait recognition.

\textbf{Skeleton Correction Network.} With information from the joint and bone feature, we use three different features as the network's input to correct the skeleton and compute the corresponding adjustment for each point: $2^{P-1}$-by-$C$ silhouette features $F_S$, $N$-by-$K$-by-$C$ skeleton feature before pooling, and the original $N$-by-$K$-by-$2$ joint or bone matrix $J$ or $B$. $F_S$ provides the sequential information to correct the joint features $F_J$ and $F_B$. $F_J$ and $F_B$ provide the framewise and feature for each node to correct the corresponding position of the joint in the frame. $J$ and $B$ provide the input order of the points to ensure the input and output order of the points are the same.

\begin{figure}[t]
    \centering
    \includegraphics[width=\linewidth]{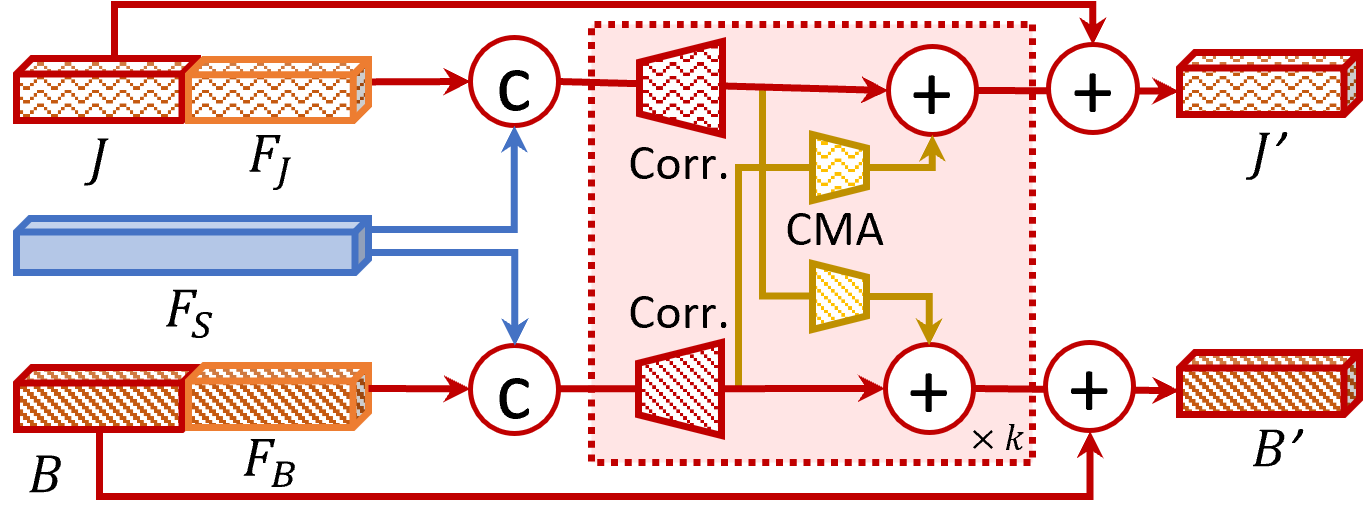}
    \caption{Architecture of the skeleton correction network. $F_J$ and $F_B$ represent the joint and bone frame-wise features encoded from $J$ (joints) and $B$ (bones), respectively. The symbol `C' denotes concatenation, and the plus sign denotes addition. `Corr' refers to the skeleton correction network, while `CMA' stands for the layer-level cross-modal adapter, and $k$ denotes the number of layers over which cross-modal skeleton correction operations are repeated between bones and joints.}
    \label{fig:correctnet}
\end{figure}

We show the architecture of the skeleton correction network in Figure~\ref{fig:correctnet}. With these three inputs, we first flatten the silhouette feature into a $2^{P-1}\times C$ vector. We then repeat it $N $-by-$ K$ times and concatenate it with the other two features to form a $N$-by-$K$-by-$(2^{P-1}\times C + C + 2)$ feature matrix.
To decode the new position $J'$ for each node in the sequence, we decode the $\Delta J$ for all the points with a reversed spatial-temporal graph convolution network to decode the $N$-by-$K$-by-$2$ adjustment for each node in $J$, and we have $J'$ for refining the individual points in $J$ following
\begin{equation}
    J' =J + \Delta J = J + SkeletonDecoder(J, F_S, F_J).
\end{equation}
The use of addition instead of directly predicting the corresponding location of the refined joints can give a relatively easier task for refinement and can preserve most of the original locations~\cite{zhu2022open}, since the original position of the joint has most of the sequential information correct and complete. By adding $\Delta J$ on $J$, we get the final refined nodes as output and process it for further encoding.\

Likewise, in the bone stream, we employ the same correction network to produce the adjustments, denoted as $\Delta B$, on the original bone matrix $B$. This refinement follows a similar procedure as in the joint stream, formulated as:
\begin{equation}
    B' = B + \Delta B = B + SkeletonDecoder(B, F_S, F_B).
\end{equation}
With this refinement, we obtain the refined bone matrix $B'$ with the assistance of encoded silhouette features, which is then utilized for further encoding in the network.

\textbf{Cross-Modal Adapter.} As the bones and joints represent the same skeleton and have connections between them, refining the skeleton and joints are also relevant to each other. We introduce the cross-modal adapter, CMA$_{i\to j}$, between these two modalities, as illustrated in Figure~\ref{fig:correctnet}, where $i$ and $j$ represent the source and target modalities. As the correction network includes multiple layers as input and their architectures are similar, for each decoded skeleton representation $F_B^{x}$ and $F_J^{x}$ at layer $x$, we employ a two-layer MLP to project the features for the other modality following
\begin{equation}
\begin{split}
    F_B^{x} &= F_B^x + CMA_{J\to B}(F_J^x)\\
    F_J^{x} &= F_J^x + CMA_{B\to J}(F_B^x).
\end{split}
\end{equation}
The refined feature $F_B^x$ will be used for the input of the next graph convolution layer $(x+1)$ to enhance inter-modal communication for more robust and accurate gait recognition.

After we get the refined skeleton $J'$ and $B'$, we apply the same skeleton feature encoder in Section~\ref{sec:gaitmix} and apply it on the refined skeleton sequence for predicting the $1$-by-$ C$ skeleton feature $F_{J'}$ and $F_{B'}$. The two skeleton feature encoders for each modality share the parameters to ensure the two embedding spaces are the same between $F_J$ and $F_{J'}$ as well as between $F_B$ and $F_{B'}$. Using the same skeleton feature encoder can also extend the data available for the encoder training to train a stabler graph convolution model the skeleton feature extraction.

With the predicted $F_{J'}$ and $F_{B'}$, we concatenate them with $2^{P-1}$-by-$C$ silhouette feature $F_S$ for representing the human body shape for \textit{GaitSTR}. In addition, we also include the joint feature before refinement as the final representation. The use of a combination of both $F_J$ and $F_{J'}$ ensures that, during training, the network can maximize its ability to distinguish the identities from the skeletons. In addition, using both of the features encoded from the two skeleton sequences gives the most representation for the task of gait recognition.

\subsection{Objectives and Inference}\label{sec:obj}
We have include two losses for training \textit{GaitSTR}: a triplet loss $L_{triplet}$ for distinguishing the same identities in the same batch and a classification loss $L_{cls}$ for the identities in training set with an MLP layer for projecting the identity feature to the number of candidates. For the combination of the two losses, we follow

\begin{equation}
    L = \lambda_1 L_{triplet} +  \lambda_2 L_{cls}
\end{equation}
and empirically set $\lambda_1$ as 1. For $\lambda_2$ we follow~\cite{lin2021gaitgl,zheng2022gait} to set it as different values for different datasets. We include further discussion and the choice of parameters in the implementation details section in Section~\ref{sec:exp}.

%% file: sections/4-exp.tex
\section{Experiments and Results}


\subsection{Experimental Details}\label{sec:exp}

\textbf{Datasets.} In our experiment, we assess our method on four public gait recognition datasets, CASIA-B~\cite{yu2006framework}, OUMVLP~\cite{he2018multi,an2020performance}, Gait3D~\cite{zheng2022gait} and GREW~\cite{zhu2021gait}. 

\textit{CASIA-B}~\cite{yu2006framework} has 124 subjects with 10 different walking variants for gait recognition. Among the 10 variants, 6 variants are for normal walking (NM), 2 variants are for the person carrying different bags (BG), and the remaining 2 variants are for different clothes (CL). 
Each subject has 110 videos captured with 10 variants from 11 different camera viewpoints distributed between 0\degree\ and 180\degree. 
We follow~\cite{chao2019gaitset,fan2020gaitpart,hou2020gln,lin2021gaitgl} and use the videos of the first 74 identities for training and the remaining 50 for inference. 
During inference, we use the first four variances in normal walking conditions (NM) to build the gallery set as the library to query test sequences.
The sequences of the remaining 2 NM variants, along with BG and CL sequences, are used as probe examples for finding the identity in the gallery.

\textit{OUMVLP}~\cite{takemura2018multi,an2020performance} is a large-scale dataset with 10,307 different identities. Each subject in this dataset has 2 different variants for normal walking (NM) conditions from 14 camera viewpoints, making 28 gait sequences. The angles of camera viewpoints are evenly distributed in two bins, 0\degree\ to 90\degree, and 180\degree\ to 270\degree. Every two neighbor viewpoints have a 15-degree gap. We follow~\cite{chao2019gaitset,fan2020gaitpart,hou2020gln,lin2021gaitgl} to use the identities with odd indexes between the 1-\textit{st} and 10,305-\textit{th} examples and build a training set with 5,153 identities. For the remaining 5,154 identities, we use the first sequence as the gallery set and the second as probes during inference.

\textit{Gait3D}~\cite{zheng2022gait} is a medium dataset compared with CASIA-B and OUMVLP for gait recognition in the wild. It includes 4,000 identities among 25,309 video sequences captured via 39 cameras. Since sequences are captured in the wild, camera positions, carried-on objects, and clothes vary from sequence to sequence. Similar to GREW~\cite{zhu2021gait}, Gait3D also provides both skeletons and silhouette sequences for each frame in the dataset. We follow~\cite{zheng2022gait} to use 3,000 identities for training and the remaining 1,000 during inference. For these 1,000 test cases, we build a probe set with 1,000 sequences for querying, as the probe set, and use the rest 5,369 sequences as the gallery set. 

\textit{GREW}~\cite{zhu2021gait} is a large in-the-wild gait recognition dataset with 128,671 sequences capturing 26,345 identities from 882 cameras. Each frame in the video has both silhouettes and poses provided. We follow~\cite{zhu2021gait} for using 20,000 identities for training and 6,000 identities as our test set. Each subject in the test set has 4 sequences, where we use two for the gallery and the other two as probe videos following the official split~\cite{zhu2021gait}. 

\textbf{Implementation Details.} For the implementation details section, we will discuss the details for the data preparation, model, and hyperparameter selection in experiments.

\textit{Data preparation.} For all four datasets, we follow OpenGait\footnote{\url{https://github.com/ShiqiYu/OpenGait}} to prepare the silhouettes for each dataset, setting the size of each frame to $64\times 44$. 
Unlike silhouettes, skeletons provided for different datasets vary. Thus, we process the joints for each dataset independently.
For the CASIA-B~\cite{yu2006framework} dataset, we follow GaitGraph~\cite{teepe2021gaitgraph} and use a pretrained HR-Net~\cite{sun2019hrnet} to generate the skeleton in MS COCO~\cite{lin2014microsoft} format with 17 joints. The number of frames used for the skeletons of CASIA-B is set to 60, using the 60 frames at the center of the entire sequence as joint input. 

For OUMVLP~\cite{takemura2018multi} dataset, we follow~\cite{an2020performance} for applying the skeletons along with the silhouette sequences, and we have skeleton sequences with 18 nodes per frame as OpenPose~\cite{cao2019openpose} format. Considering that the sequence length in OUMVLP is shorter than CASIA-B, we set the fixed frame number to 25 for each sequence. For videos shorter than 25, we repeat the frames until we have 25 frames. 

For Gait3D~\cite{zheng2022gait} and GREW~\cite{zhu2021gait}, since skeletons are collected in the wild, we normalize each skeleton by setting their height to 2 and move their center to the origin point $(0, 0)$. This can ensure that the position of the skeletons is aligned across different videos and will not change significantly.

In addition to joints, we generate bones based on predefined neighbor link relationships between joints, represented as directional vectors calculated from the differences in coordinates between linked joints. Compared to the joints in the skeletons, the connected bones are defined by neighbor link relationships, emphasizing the numerical connectivity that is not explicitly captured in ST-GCN~\cite{yan2018spatial}.

\input{tables/casiab}

\textit{Network details.} In our network, we have two different encoders. For our silhouette feature encoder, we follow GaitGL~\cite{lin2021gaitgl} to build the encoder for CASIA-B. For Gait3D, OUMVLP, and GREW, we follow GaitBase~\cite{fan2023opengait} to encode silhouette features.  Note that for GaitMix and GaitRef, we follow~\cite{zhu2023gaitref} to use OpenGait~\cite{fan2023opengait} for Gait3D and GaitGL~\cite{lin2021gaitgl} for GREW respectively. For the skeleton feature encoder, we follow ST-GCN~\cite{yan2018spatial} for encoding the skeletons into the same embedding dimension $N_{out}$ as the silhouette feature encoder. The dimension of the hidden layers of ST-GCN is set to [64, 64, 128, 128, $n_{out}$]. The two skeleton decoders of the \textit{GaitSTR} both use the reversed shape of the ST-GCN, with [128, 64, 64, 3] as the hidden dimensions and the number of CMA, $K$ is set to 3. For the encoder and decoder network, we have compared ST-GCN along with other choices, such as MS-G3D~\cite{liu2020disentangling} for ablation study.

\textit{Model training.} In our model, we follow~\cite{lin2021gaitgl,fan2023opengait} for choosing the hyperparameters. For CASIA-B, we use an Adam optimizer~\cite{kingma2014adam} with $10^{-4}$ as the learning rate for 80,000 iterations. We decay the learning rate once at 70,000 iterations for CASIA-B as $\frac{1}{10}$ of its original value. For Gait3D, OUMVLP and GREW, we use the SGD optimizer for 60,000, 120,000 and 180,000 iterations, respectively and set the initial learning rate as $10^{-3}$. The learning rate is decayed to $\frac{1}{10}$ three times for these three datasets, at iteration 20,000, 40,000 and 50,000 for Gait3D, 60,000, 80,000 and 100,000 for OUMVLP, and 80,000, 120,000 and 150,000 for GREW. For all four datasets we use, we follow~\cite{lin2021gaitgl,fan2023opengait} for using 1 for both $\lambda_1$ and $\lambda_2$ following our ablation results.

\textit{Metrics and evaluations.} During inference, for each example in the probe set, we use $L_2$ similarity to find the nearest example in the gallery set. For CASIA-B and OUMVLP, we evaluate the top-1 accuracy for the prediction. For GREW, we evaluate top-1, 5, 10 and 20 accuracies. For Gait3D, we assess top-1 and top-5 accuracies along with mAP and mINP following~\cite{ye2021deep} for assessing since all the correct matches should have low-rank values when pairing the probe example with correct identities in the gallery.

For baseline methods, we compare with state-of-the-art gait recognition approaches, including GaitNet~\cite{song2019gaitnet}, GaitSet~\cite{chao2019gaitset}, GaitPart~\cite{fan2020gaitpart}, GLN~\cite{hou2020gln}, GaitGL~\cite{lin2021gaitgl}, ModelGait~\cite{li2020end}, and CSTL~\cite{huang2021context}. Additionally, we compare with PoseGait~\cite{liao2020model} and GaitGraph~\cite{teepe2021gaitgraph}, which utilize skeleton sequences as their input. For baseline comparison, we include GaitMix and GaitRef~\cite{zhu2023gaitref}, which use simple concatenation as described in Section~\ref{sec:gaitmix} and only utilize the skeleton correction network as outlined in Section~\ref{sec:gaitref}, respectively. \textit{GaitSTR}, along with these methods, employs 2-D convolution for binarized silhouette feature extraction. We also include comparisons with other methods which use different modalities, such as GaitEdge~\cite{liang2022gaitedge} and MvModelGait~\cite{li2021end}.

\subsection{Results and Analysis}\label{sec:res}
In this subsection, we first present the numerical results for CASIA-B~\cite{yu2006framework}, OUMVLP~\cite{he2018multi,an2020performance}, Gait3D~\cite{zheng2022gait}, and GREW~\cite{zhu2021gait} compared to other state-of-the-art methods. We then delineate and analyze the enhancements conferred by the multi-modal gait model as opposed to the refinement of the skeletons from the silhouettes. In addition, we compare the use of skeletons with 3-D body shapes on the top of silhouettes for gait recognition. Finally, we present an ablation study for our model and visualizations that illustrate the corrected skeletons informed by silhouette guidance.

\input{tables/oumvlp}

\textbf{Numerical Results.} We present our numerical performance on the four datasets used in our experiments in Table~\ref{tab:casiab-1}, \ref{tab:oumvlp}, \ref{table:gait3d}, and \ref{table:grew}, respectively. We follow the official splits of these four datasets for gallery and probe constructions. For CASIA-B and OUMVLP, identical-view cases are excluded.

\input{tables/gait3d}

For all four datasets evaluated, we outperform the existing state-of-the-art methods with \textit{GaitSTR}. In Table~\ref{tab:casiab-1}, on CASIA-B, we achieve the best performance on all splits. Specifically, on NM, BG, and CL, we reduce the error rates from $2.1\%$, $5.6\%$, and $15.8\%$ to $1.6\%$, $3.8\%$, and $10.4\%$, respectively, which correspond to a relative reduction of $23.8\%$, $32.1\%$, and $34.2\%$ in error rates compared with the best model using 2-D convolution for silhouette feature extraction, while we also show similar performance compared to methods~\cite{sun2023trigait} using 3-D convolution. The margin of improvement is even greater for NM and CL settings when compared to our baseline silhouette encoder, GaitGL, where we demonstrate a $33.3\%$ and $37.0\%$ relative reduction for the average rank-1 predictions across all camera views. Furthermore, when compared with GaitEdge~\cite{liang2022gaitedge} and MvModelGait~\cite{li2021end}, which utilize RGB images and viewpoint angles not typically present in public datasets, \textit{GaitSTR} still exhibits superior performance, indicating the effectiveness of using the skeletons along with silhouettes is able to outperform the methods directly using the RGB images as input, which actually include all the information stored in the silhoeuttes and skeletons.

For the other three datasets, on OUMVLP in Table~\ref{tab:oumvlp}, we show small improvement compared with GaitBase~\cite{fan2023opengait} for the top-1 accuracy, while we outperform it along with other methods for all the metrics on Gait3D~\cite{zheng2022gait} and GREW~\cite{zhu2021gait} in Table~\ref{table:gait3d} and \ref{table:grew}, which we show $3.1\%$ and $3.9\%$ improvements on rank-1 accuracies respectively compared with other state-of-the-art methods using 2-D convolution as silhouette feature encoders. We also show improvement on other metrics, such as mAP and mINP~\cite{zheng2022gait} used by Gait3D~\cite{zheng2022gait}.

\textbf{Improvement of Skeleton Refinement.} With the inclusion of both bones and joints to represent skeletons, we further analyze the improvement from introducing a new modality and the manner in which these modalities are utilized in \textit{GaitSTR}. We present the results in Table~\ref{table:mixrefstr}, showing rank-1 accuracy on CASIA-B~\cite{yu2006framework}. We begin with our baseline method, GaitGL~\cite{lin2021gaitgl}, which operates on a single modality, and then proceed to analyze the introduction of joints and bones, as well as various ways of integrating them with silhouettes for different combinations.
As our model includes an extra refinement model for skeleton correction, we also include the corresponding number of parameters used in the network. Even with the increase in the number of parameters required, the inference time for a 30-frame gait sequence is about 45$\times$ faster than real-time on a single Nvidia 3090 Ti GPU.

\input{tables/grew}
\input{tables/mixrefstr}

When comparing the use of silhouettes as the only input for gait recognition, the introduction of joints as skeletons displays an improvement for both simple feature correction, as seen with GaitMix~\cite{zhu2023gaitref}, and the use of silhouettes to refine joints, as with GaitRef~\cite{zhu2023gaitref} across all three metrics. GaitRef, which uses silhouettes to refine the joints, provides better recognition accuracy compared to simply aggregating the two features.

Furthermore, the introduction of bones in both GaitMix and GaitRef leads to additional improvements over the use of joints alone with silhouettes. For GaitRef, we treat the newly introduced bones similarly to joints and utilize silhouette features to refine them. We then concatenate the three encoded features (silhouettes, joints, and bones) for recognition. Including a single-sided bone-to-joint cross-modal adapter results in a slight decrease in performance, whereas incorporating adapters on both sides, as in \textit{GaitSTR}, shows better performance. The refinement from one side creates inconsistencies across the two skeletal representations, while the two-way sequential refinement provides consistent enhancement across these modalities.

\textbf{Skeleton and Body Shape for Gait Recognition.} In addition to silhouettes and skeletons, Gait3D~\cite{zheng2022gait} provides 3-D body shapes alongside silhouette sequences, which are utilized by SMPLGait~\cite{zheng2022gait}. In Table~\ref{table:gait3d}, we provide a comparison of using 3-D body shapes as in SMPLGait~\cite{zheng2022gait} and using skeletons by removing the skeleton correction network and cross-modal adapters, which directly aggregate skeleton features with joint features as in GaitMix~\cite{zhu2023gaitref} and GaitRef~\cite{zhu2023gaitref}. For a fair comparison, we use GaitMix and GaitRef~\cite{zhu2023gaitref} with the OpenGait~\cite{fan2023opengait} baseline without augmentation, as SMPLGait~\cite{zhu2021gait} is not implemented on the latest OpenGait configuration with data augmentation. 

Compared to using silhouettes as the only input modality, the inclusion of skeletons and body shapes both enhance recognition accuracy. In SMPLGait, skeleton information is partially integrated into the generated 3-D body shapes for gait recognition, making SMPLGait yields similar performance as GaitMix~\cite{zhu2023gaitref}, which includes joints and bones, across all four metrics.

When compared to SMPLGait, which uses a 3-D body shape as the second modality, GaitRef~\cite{zhu2023gaitref} with bone inputs for the refined skeletons achieves better recognition performance. Considering that the generation of SMPL body shapes also requires skeletons~\cite{sun2022putting}, inaccurate pose estimation in 3-D body shape generation can hinder the model's ability to correctly interpret noisy body shapes with erroneous poses in SMPLGait~\cite{zheng2022gait}. GaitRef, however, does not suffer from this issue with refined skeletons.

\input{tables/abla-combination}
\input{tables/abla-feat}

\textbf{Ablation Studies.} For ablation studies, we present results on: 1) different methods of combining skeleton and silhouette features, 2) various skeleton encoder and decoder networks in comparison with other skeleton refinement methods, and 3) the inputs of the skeleton correction network. All experiments were conducted on the CASIA-B dataset~\cite{yu2006framework} for each of the three different settings, and we present the Top-1 accuracy for the final gait recognition results. The results are shown in Tables~\ref{table:abla-comb}, \ref{table:abla-feat}, and \ref{tab:pose}, respectively. Since the joint and bones branches are identical and exhibit similar performance, our ablation experiments focus on the joints branch as skeleton representation.

\textbf{\textit{(i) Feature Combination.}} In addition to concatenating the features, we also repeat and pad the skeleton feature along with each segment of the silhouette features to provide the guidance for different level of silhouette embeddings, which we label as `padding'. We show the results in Table~\ref{table:abla-comb}. For comparison, we also add the performance of GaitGL~\cite{lin2021gaitgl} in the table, which only uses the silhouette feature for gait recognition and is our backbone baseline on CASIA-B. We observe that padding the skeleton feature alongside each size of the silhouette feature results in worse performance compared to concatenating the refined feature just once as the final recognition feature. Padding the skeleton feature multiple times may cause the skeleton input to dominate the feature space, whereas concatenating it once allows the silhouette features to contribute robust information about the pose and remain less sensitive to skeletons.

\textbf{\textit{(ii) Encoder-decoder Variations.}} For the choice of the skeleton encoder and skeleton correction decoder, we select between two state-of-the-art skeleton action recognition models: ST-GCN~\cite{yan2018spatial} and MS-G3D~\cite{liu2020disentangling}. The results are presented in Table~\ref{table:abla-feat}. Following previous observations, where using joints as the skeleton representation follows the same trend as using both joints and bones, we opt for using only joints as the skeleton representation for this ablation. Both MS-G3D and ST-GCN show improvement in performance. However, in our experiments, MS-G3D requires significantly more GPU memory and at least double the training time for each module introduced. Considering their comparable performance and time efficiency, we choose ST-GCN for both the encoder and decoder in our final pipeline.
\input{tables/ablasupp}

\textbf{\textit{(iii) Skeleton Refinement.}} For skeleton refinement, we compare the refining of the skeleton sequence using silhouettes with neighbor smoothing (average and Gaussian window for the neighboring three frames) and SmoothNet~\cite{zeng2021smoothnet} (pretrained on H36m~\cite{ionescu2013human3}) on the CASIA-B dataset based on Top-1 accuracy. The results, displayed in Table~\ref{table:abla-feat}, demonstrate that refining skeleton with silhouettes outperforms the other methods. Among the three variations, 3-frame Gaussian smoothing shows a slight improvement but still falls short compared to using silhouettes.

Different from naive temporal smoothing, which can result in poses inconsistent in the sequence, integrating silhouette features introduces walking patterns not present in the skeletons, aiding their self-refinement for gait recognition. Compared to skeletons refined from the skeleton sequence alone, external knowledge from encoded silhouette embeddings reduces ambiguity, providing ID-specific information during training when the walking pattern cannot be correctly extracted from the skeleton alone.

\textbf{\textit{(iv) Input of the Skeleton Correction Network.}} Considering three distinct inputs in our skeleton correction network, $F_J$, $F_S$, and $J$, we investigate each component's contributions and present the results in Table~\ref{tab:pose} using three splits of the CASIA-B datasets. We note that when either of the three input is excluded, there is a significant drop in performance. The skeleton correction network capitalizes on temporal consistency in the skeleton sequences for correction, while the additional silhouette information provides external support for an enhanced understanding. 

\textbf{\textit{(v) Weights for losses and number of frames.}} As we set $\lambda_1$ and $\lambda_2$ to be 1 in our main experiments, we include ablation results for different $\lambda$ combinations on CASIA-B in Table~\ref{table:abla-supp}, along with the number of frames of videos used in our experiment. We note that setting both $\lambda_1$ and $\lambda_2$ to 1 shows the best average performance. In addition, with more frames available, models are able to perform better compared with fewer frames available.

\begin{figure}[t]
    \centering
    \includegraphics[width=\linewidth]{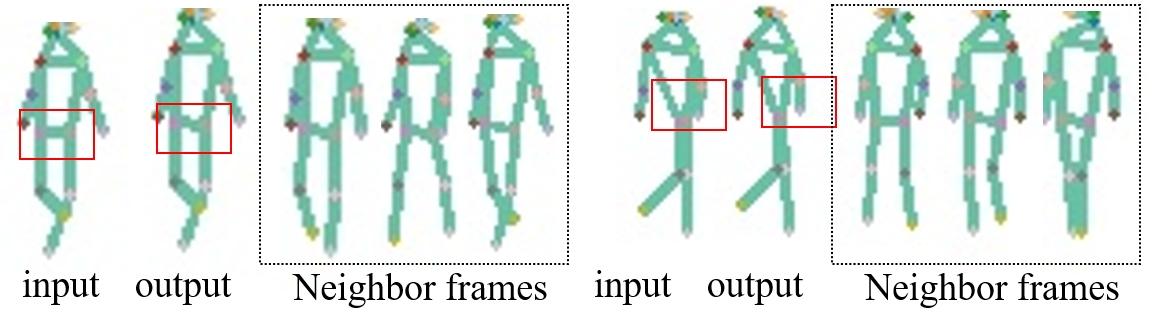}
    \caption{Visualization of successful and failed refined skeletons with \textit{GaitSTR}. For each example, from left to right, we have original skeletons, refined skeletons and its neighbor frames.}
    \label{fig:vis}
\end{figure}

\textbf{Skeletons Visualization.} We present two examples from \textit{GaitSTR} compared to the original skeletons in Figure~\ref{fig:vis}, accompanied by the three nearest silhouettes from a similar timestamp. With two modalities of representations for gait, \textit{GaitSTR} can make more precise modifications to the skeletons' nodes. However, it still fails on some obvious errors, as seen in the second example in Figure~\ref{fig:vis}.

%% file: tables/casiab.tex
\begin{table*}[t]
\caption{Gait recognition results on CASIA-B dataset, excluding identical-view cases. GaitEdge* requires RGB frames and uses the re-segmented CASIA-B* silhouettes instead of CASIA-B, and MvModelGait requires the input camera viewpoints. TriGait includes a 3-D convolution feature extractor which requires much heavier computation than the 2-D encoders used by other methods in the table. We mark the best results among all the methods in bold and the best results in our baseline methods with underline.
} 
\label{tab:casiab-1}
\centering
\def\lw{0.6}
\def\la{2}
\def\ls{0.0}
\resizebox{0.92\linewidth}{!}
{
\begin{tabu}{p{1.3cm}p{2.4cm}p{\ls cm}p{\lw cm}<{\centering}p{\lw cm}<{\centering}p{\lw cm}<{\centering}p{\lw cm}<{\centering}p{\lw cm}<{\centering}p{\lw cm}<{\centering}p{\lw cm}<{\centering}p{\lw cm}<{\centering}p{\lw cm}<{\centering}p{\lw cm}<{\centering}p{\lw cm}<{\centering}p{\ls cm}p{\lw cm}<{\centering}} \toprule
\multirow{2}{*}{Probe}&\multirow{2}{*}{Method} && \multicolumn{11}{c}{Camera Positions} && \multirow{2}{*}{Mean} \\
 
 \cline{4-14}  \\ [-8pt]
            &&& 0\degree   & 18\degree  & 36\degree  & 54\degree  & 72\degree  & 90\degree 
            &   108\degree & 126\degree & 144\degree & 162\degree & 180\degree && \\
\midrule
\multirow{15}{*}{NM \#5-6}
         & PoseGait \cite{liao2020model}        && 55.3 & 69.6 & 73.9 & 75.0 & 68.0 & 68.2 & 71.1 & 72.9 & 76.1 & 70.4 & 55.4 && 68.7\\
         & GaitNet \cite{song2019gaitnet}       && 91.2 & 92.0 & 90.5 & 95.6 & 86.9 & 92.6 & 93.5 & 96.0 & 90.9 & 88.8 & 89.0 && 91.6\\
         & GaitGraph \cite{teepe2021gaitgraph}  && 85.3 & 88.5 & 91.0 & 92.5 & 87.2 & 86.5 & 88.4 & 89.2 & 87.9 & 85.9 & 81.9 && 87.7\\
         & GaitSet \cite{chao2019gaitset}       && 91.1 & 98.0 & 99.6 & 97.8 & 95.4 & 93.8 & 95.7 & 97.5 & 98.1 & 97.0 & 88.2 && 95.6\\
         & GaitPart \cite{fan2020gaitpart}      && 94.0 & 98.7 & 99.3 & 98.8 & 94.8 & 92.6 & 96.4 & 98.3 & 99.0 & 97.4 & 91.2 && 96.4\\
         & GLN \cite{hou2020gln}                && 93.8 & 98.5 & 99.2 & 98.0 & 95.2 & 92.9 & 95.4 & 98.5 & 99.0 & 99.2 & 91.9 && 96.5\\
         & GaitGL \cite{lin2021gaitgl}          && 95.3 & 97.9 & 99.0 & 97.8 & 96.1 & 95.3 & 97.2 & 98.9 & 99.4 & 98.8 & 94.5 && 97.3\\
         & CSTL \cite{huang2021context}         && 97.2 & 99.0 & 99.2 & 98.1 & 96.2 & 95.5 & 97.7 & 98.7 & 99.2 & 98.9 & 96.5 && 97.8\\
         & ModelGait \cite{li2020end}           && 96.9 & 97.1 & 98.5 & 98.4 & 97.7 & 98.2 & 97.6 & 97.6 & 98.0 & 98.4 & 98.6 && \underline{97.9}\\
\cline{2-16}  \\ [-8pt]
         & GaitMix \cite{zhu2023gaitref}        && 96.6 & 98.6 & 99.2 & 98.0 & 97.1 & 96.2 & 97.5 & 98.9 & 99.3 & 99.0 & 94.7 && 97.7 \\
         & GaitRef  \cite{zhu2023gaitref}       && 97.2 & 98.7 & 99.1 & 98.0 & 97.3 & 97.0 & 98.0 & 99.4 & 99.4 & 98.9 & 96.4 && 98.1 \\
         & GaitSTR                              && 97.2 & 98.4 & 99.2 & 98.3 & 97.6 & 97.8 & 97.9 & 99.3 & 99.3 & 99.3 & 97.6 && \textbf{98.4} \\     
\cline{2-16}  \\ [-8pt]
         \rowfont{\protect\leavevmode\color{gray!60}}& MvModelGait \cite{li2021end}         && 97.5 & 97.6 & 98.6 & 98.8 & 97.7 & 98.9 & 98.9 & 97.3 & 97.6 & 97.8 & 97.9 &&  {98.1}\\
         \rowfont{\protect\leavevmode\color{gray!60}}& {GaitEdge* \cite{liang2022gaitedge}}    && {97.2} & {99.1} & {99.2} & {98.3} & {97.3} & {95.5} & {97.1} & {99.4} & {99.3} & {98.5} & {96.4} && {97.9}\\
         \rowfont{\protect\leavevmode\color{gray!60}}& TriGait \cite{sun2023trigait}         && 97.0 & 98.6 & 98.3 & 98.3 & 98.4 & 97.0 & 98.6 & 99.0 & 98.9 & 98.4 & 97.4 && 98.2\\
\midrule
\multirow{15}{*}{BG \#1-2}
         & PoseGait \cite{liao2020model}        && 35.3 & 47.2 & 52.4 & 46.9 & 45.5 & 43.9 & 46.1 & 48.1 & 49.4 & 43.6 & 31.1 && 44.5\\
         & GaitNet \cite{song2019gaitnet}       && 83.0 & 87.8 & 88.3 & 93.3 & 82.6 & 74.8 & 89.5 & 91.0 & 86.1 & 81.2 & 85.6 && 85.7\\
         & GaitGraph \cite{teepe2021gaitgraph}  && 75.8 & 76.7 & 75.9 & 76.1 & 71.4 & 73.9 & 78.0 & 74.7 & 75.4 & 75.4 & 69.2 && 74.8\\
         & GaitSet \cite{chao2019gaitset}       && 87.0 & 93.8 & 94.6 & 92.9 & 88.2 & 83.0 & 86.6 & 92.6 & 95.7 & 92.9 & 83.4 && 90.1\\
         & GaitPart \cite{fan2020gaitpart}      && 89.5 & 94.5 & 95.3 & 93.5 & 88.5 & 83.9 & 89.0 & 93.6 & 96.0 & 94.1 & 85.3 && 91.2\\
         & GLN \cite{hou2020gln}                && 92.2 & 95.6 & 96.7 & 94.3 & 91.8 & 87.8 & 91.4 & 95.1 & 96.3 & 95.7 & 87.2 && 93.1\\
         & GaitGL \cite{lin2021gaitgl}          && 93.0 & 95.7 & 97.0 & 95.9 & 93.3 & 90.0 & 91.9 & 96.8 & 97.5 & 96.9 & 90.7 && \underline{94.4}\\
         & CSTL \cite{huang2021context}         && 91.7 & 96.5 & 97.0 & 95.4 & 90.9 & 88.0 & 91.5 & 95.8 & 97.0 & 95.5 & 90.3 && 93.6\\
         & ModelGait \cite{li2020end}           && 94.8 & 92.9 & 93.8 & 94.5 & 93.1 & 92.6 & 94.0 & 94.5 & 89.7 & 93.6 & 90.4 && 93.1\\
\cline{2-16}  \\ [-8pt]
         & GaitMix  \cite{zhu2023gaitref}       && 94.4 & 96.7 & 96.8 & 96.1 & 94.3 & 90.4 & 93.5 & 97.4 & 98.0 & 97.2 & 92.2 && 95.2\\
         & GaitRef  \cite{zhu2023gaitref}       && 94.4 & 96.4 & 97.3 & 96.8 & 96.2 & 92.2 & 94.4 & 97.2 & 98.7 & 97.9 & 93.3 && 95.9\\
         & GaitSTR                              && 95.3 & 97.1 & 97.8 & 96.8 & 96.1 & 93.2 & 94.3 & 96.8 & 98.3 & 98.3 & 94.0 && \textbf{96.2} \\      
\cline{2-16}  \\ [-8pt]
         \rowfont{\protect\leavevmode\color{gray!60}}& MvModelGait \cite{li2021end}         && 93.9 & 92.5 & 92.9 & 94.1 & 93.4 & 93.4 & 95.0 & 94.7 & 92.9 & 93.1 & 92.1 && 93.4\\
         \rowfont{\protect\leavevmode\color{gray!60}}& GaitEdge* \cite{liang2022gaitedge}    && 95.3 & 97.4 & 98.4 & 97.6 & 94.3 & 90.6 & 93.1 & 97.8 & 99.1 & 98.0 & 95.0 && {96.1}\\
         \rowfont{\protect\leavevmode\color{gray!60}}& TriGait \cite{sun2023trigait}         && 91.8 & 94.3 & 95.2 & 96.6 & 96.5 & 93.7 & 95.9 & 97.6 & 97.4 & 96.9 & 93.8 && 95.4\\ 
\midrule
\multirow{15}{*}{CL \#1-2}
         & PoseGait \cite{liao2020model}        && 24.3 & 29.7 & 41.3 & 38.8 & 38.2 & 38.5 & 41.6 & 44.9 & 42.2 & 33.4 & 22.5 && 36.0\\
         & GaitNet \cite{song2019gaitnet}       && 42.1 & 58.2 & 65.1 & 70.7 & 68.0 & 70.6 & 65.3 & 69.4 & 51.5 & 50.1 & 36.6 && 58.9\\
         & GaitGraph \cite{teepe2021gaitgraph}  && 69.6 & 66.1 & 68.8 & 67.2 & 64.5 & 62.0 & 69.5 & 65.6 & 65.7 & 66.1 & 64.3 && 66.3\\
         & GaitSet \cite{chao2019gaitset}       && 71.0 & 82.6 & 84.0 & 80.0 & 71.7 & 69.1 & 72.1 & 76.7 & 78.5 & 77.2 & 63.4 && 75.1\\
         & GaitPart \cite{fan2020gaitpart}      && 72.5 & 82.8 & 86.0 & 82.2 & 79.5 & 71.0 & 77.7 & 80.8 & 82.9 & 81.4 & 67.7 && 78.6\\
         & GLN \cite{hou2020gln}                && 78.5 & 90.4 & 90.3 & 85.1 & 80.2 & 75.8 & 78.1 & 81.8 & 80.9 & 83.2 & 72.6 && 81.5\\
         & GaitGL \cite{lin2021gaitgl}          && 71.7 & 90.5 & 92.4 & 89.4 & 84.9 & 78.1 & 83.1 & 87.5 & 89.1 & 83.9 & 67.4 && 83.5\\
         & CSTL \cite{huang2021context}         && 78.1 & 89.4 & 91.6 & 86.6 & 82.1 & 79.9 & 81.8 & 86.3 & 88.7 & 86.6 & 75.3 && \underline{84.2}\\
         & ModelGait \cite{li2020end}           && 78.2 & 81.0 & 82.1 & 82.8 & 80.3 & 76.9 & 75.5 & 77.4 & 72.3 & 73.5 & 74.2 && 77.6\\
\cline{2-16}  \\ [-8pt]
         & GaitMix   \cite{zhu2023gaitref}      && 79.2 & 89.5 & 94.2 & 90.0 & 84.9 & 80.3 & 85.2 & 89.2 & 90.3 & 86.9 & 73.7 && 85.8\\
         & GaitRef    \cite{zhu2023gaitref}     && 81.4 & 93.3 & 94.3 & 91.6 & 87.8 & 83.9 & 88.5 & 91.7 & 91.6 & 89.1 & 75.0 && 88.0\\
         & GaitSTR                              && 83.8 & 94.0 & 94.9 & 94.3 & 90.7 & 85.5 & 89.2 & 91.8 & 92.8 & 90.7 & 78.2 && \textbf{89.6} \\    
\cline{2-16}  \\ [-8pt]
         \rowfont{\protect\leavevmode\color{gray!60}}& MvModelGait \cite{li2021end}         && 77.0 & 80.0 & 83.5 & 86.1 & 84.5 & 84.9 & 80.6 & 80.4 & 77.4 & 76.6 & 76.9 && 80.7\\
         \rowfont{\protect\leavevmode\color{gray!60}}& GaitEdge* \cite{liang2022gaitedge}    && 84.3 & 92.8 & 94.3 & 92.2 & 84.6 & 83.0 & 83.0 & 87.5 & 87.4 & 85.9 & 75.0 && {86.4}\\
         \rowfont{\protect\leavevmode\color{gray!60}}& TriGait \cite{sun2023trigait}         && 91.7 & 93.2 & 96.9 & 97.0 & 95.2 & 94.0 & 94.6 & 95.3 & 94.1 & 94.1 & 90.8 && 94.3\\
\bottomrule
\end{tabu}
}
\end{table*}

%% file: tables/oumvlp.tex
\begin{table*}[tb]
\caption{Gait recognition results for accuracy across all the test views on OUMVLP dataset, excluding identical-view cases. } 
\label{tab:oumvlp}
\centering
\def\lw{0.7}
\def\ls{0.06}
\resizebox{0.97\linewidth}{!}
{
\begin{tabular}{p{2.7cm}p{\ls cm}p{\lw cm}<{\centering}p{\lw cm}<{\centering}p{\lw cm}<{\centering}p{\lw cm}<{\centering}p{\lw cm}<{\centering}p{\lw cm}<{\centering}p{\lw cm}<{\centering}p{\lw cm}<{\centering}p{\lw cm}<{\centering}p{\lw cm}<{\centering}p{\lw cm}<{\centering}p{\lw cm}<{\centering}p{\lw cm}<{\centering}p{\lw cm}<{\centering}p{\ls cm}p{0.9 cm}<{\centering}} \toprule
\multirow{2}{*}{Method} && \multicolumn{14}{c}{Camera Positions} && \multirow{2}{*}{Mean} \\

 \cline{3-16}  \\ [-8pt]
            && 0\degree   & 15\degree  & 30\degree  & 45\degree  & 60\degree  & 75\degree  & 90\degree 
            &   180\degree & 195\degree & 210\degree & 225\degree & 240\degree & 255\degree & 270\degree && \\
\midrule
GEINet \cite{shiraga2016geinet} && 23.2 & 38.1 & 48.0 & 51.8 & 47.5 & 48.1 & 43.8 & 27.3 & 37.9 & 46.8 & 49.9 & 45.9 & 45.7 & 41.0 && 42.5\\
GaitSet \cite{chao2019gaitset}  && 79.2 & 87.7 & 89.9 & 90.1 & 87.9 & 88.6 & 87.7 & 81.7 & 86.4 & 89.0 & 89.2 & 87.2 & 87.7 & 86.2 && 87.0\\
GaitPart \cite{fan2020gaitpart} && 82.8 & 89.2 & 90.9 & 91.0 & 89.7 & 89.9 & 89.3 & 85.1 & 87.7 & 90.0 & 90.1 & 89.0 & 89.0 & 88.1 && 88.7\\
GLN \cite{hou2020gln}           && 83.8 & 90.0 & 91.0 & 91.2 & 90.3 & 90.0 & 89.4 & 85.3 & 89.1 & 90.5 & 90.6 & 89.6 & 89.3 & 88.5 && 89.2\\
GaitGL \cite{lin2021gaitgl}     && 84.2 & 89.8 & 91.3 & 91.7 & 90.8 & 91.0 & 90.4 & 88.1 & 88.2 & 90.5 & 90.5 & 89.5 & 89.7 & 88.8 && 89.6\\
MvModelGait \cite{li2021end}    && 87.7 & 89.7 & 91.1 & 90.1 & 89.8 & 90.3 & 90.3 & 88.1 & 89.4 & 89.4 & 90.0 & 90.8 & 90.0 & 89.7 && 89.7\\
CSTL \cite{huang2021context}    && 87.1 & 91.0 & 91.5 & 91.8 & 90.6 & 90.8 & 90.6 & 89.4 & 90.2 & 90.5 & 90.7 & 89.8 & 90.0 & 89.4 && 90.2\\
GaitBase \cite{fan2023opengait} && 87.2 & 91.2 & 91.8 & 92.0 & 91.4 & 91.2 & 90.8 & 88.9 & 90.4 & 91.1 & 91.3 & 90.7 & 90.5 & 90.0 && \underline{90.6}\\
\midrule
GaitMix~\cite{zhu2023gaitref} && 85.4 & 90.3 & 91.2 & 91.5 & 91.2 & 90.9 & 90.5 & 88.9 & 88.7 & 90.3 & 90.5 & 89.8 & 89.6 & 88.9 && {89.9}\\
GaitRef~\cite{zhu2023gaitref} && 85.7 & 90.5 & 91.6 & 91.9 & 91.3 & 91.3 & 90.9 & 89.3 & 89.0 & 90.8 & 90.8 & 90.1 & 90.1 & 89.5 && {90.2}\\
GaitSTR &&                         87.6 & 91.5 & 91.8 & 92.1 & 91.5 & 91.3 & 91.0 & 89.2 & 90.7 & 91.1 & 91.3 & 90.8 & 90.6 & 90.2 && \textbf{90.8}\\
\bottomrule
\end{tabular}
}
\end{table*}

%% file: tables/gait3d.tex
\begin{table}
\begin{center}
\caption{Gait recognition results reported on the Gait3D dataset with $64\times 44$ as input sizes. For all four metrics, higher values of the same metric indicate better performance. B represents the bone input.}
\label{table:gait3d}
{
\begin{tabu}{p{2.4cm}p{1.cm}<{\centering}p{1.cm}<{\centering}p{1.cm}<{\centering}p{1.cm}<{\centering}}
\toprule
Methods &  Rank@1 & Rank@5 & mAP & mINP\\
 \midrule
GaitSet \cite{chao2019gaitset}&     36.70 & 58.30 & 30.01 &  17.30\\
GaitPart \cite{fan2020gaitpart} &   28.20 & 47.60 & 21.58 &  12.36\\
GLN \cite{hou2020gln} &             31.40 & 52.90 & 24.74 &  13.58\\
GaitGL \cite{lin2021gaitgl} &       29.70 & 48.50 & 22.29 &  13.26\\
OpenGait \cite{zheng2022gait} &     42.90 & 63.90 & 35.19 &  20.83\\
CSTL \cite{huang2021context} &      11.70 & 19.20 & 5.59 & 2.59 \\
GaitBase \cite{fan2023opengait} & \underline{62.00} & \underline{78.80} & \underline{53.17} & \underline{35.33}\\
\midrule
SMPLGait \cite{zheng2022gait} &     {46.30} & {64.50} & {37.16} & {22.23}\\
GaitMix \cite{zhu2023gaitref}   & 46.20 & 66.20 & 37.08 & 22.85\\
GaitRef \cite{zhu2023gaitref}  & 49.00 & 69.30 & 40.69 & 25.26 \\
\midrule
GaitSTR & \textbf{65.10} & \textbf{81.30} & \textbf{55.59} & \textbf{36.84}    \\
\bottomrule
\end{tabu}}
\end{center}
\end{table}

%% file: tables/grew.tex
\begin{table}
\begin{center}
\caption{Rank-1, 5, 10 and 20 accuracies on GREW dataset.}
\label{table:grew}
\def\lw{1.}
{
\begin{tabular}{p{2.cm}p{\lw cm}<{\centering}p{\lw cm}<{\centering}p{\lw cm}<{\centering}p{\lw cm}<{\centering}}
\toprule
Methods &  Rank-1 & Rank-5 & Rank-10 & Rank-20  \\
 \midrule
PoseGait \cite{liao2020model} & 0.2 & 1.1 & 2.2 & 4.3\\
GaitGraph \cite{teepe2021gaitgraph} & 1.3 & 3.5 & 5.1 & 7.5\\
GEINet \cite{shiraga2016geinet} & 6.8 & 13.4 & 17.0 & 21.0\\
TS-CNN \cite{wu2016comprehensive} & 13.6 & 24.6 & 30.2 & 37.0\\
GaitSet \cite{chao2019gaitset}& 46.3 & 63.6 & 70.3 & 76.8\\
GaitPart \cite{fan2020gaitpart} & 44.0 & 60.7 & 67.4 & 73.5 \\
CSTL \cite{huang2021context} & 50.6 & 65.9 & 71.9 & 76.9 \\
GaitGL \cite{lin2021gaitgl} & {51.4} & {67.5} & {72.8} & {77.3} \\
GaitBase \cite{fan2023opengait} & \underline{60.1} & \underline{75.5} & \underline{80.4} & \underline{84.1}\\
\midrule
GaitMix~\cite{zhu2023gaitref} & {52.4} & 67.4 & {72.9} & 77.2 \\
GaitRef~\cite{zhu2023gaitref} & {53.0} & {67.9} & {73.0} & {77.5}\\
GaitSTR   & \textbf{64.0} & \textbf{78.5} & \textbf{83.2} & \textbf{86.3}   \\
\bottomrule
\end{tabular}}
\end{center}
\end{table}

%% file: tables/mixrefstr.tex
\begin{table}
\caption{Rank-1 accuracy of the variations skeletons in addition to silhouettes for gait recognition on CASIA-B. `Sils.' represents silhouettes. }
\label{table:mixrefstr}
\begin{center}
\resizebox{\columnwidth}{!}
{
\begin{tabular}{p{2cm}p{2.2cm}p{.6cm}<{\centering}p{.6cm}<{\centering}p{.6cm}<{\centering}p{1cm}<{\centering}}
\toprule
Input Modality & \multicolumn{1}{c}{Methods}  & NM & CL & BG  & Params \\
\midrule
Sil. only & GaitGL~\cite{lin2021gaitgl} & 97.3 & 94.4 & 83.5 & 3.82M\\
\midrule
\multirow{2}{*}{Sil. + Joint}  
&GaitMix~\cite{zhu2023gaitref} & 97.7 & 95.2 & 85.8 & 3.84M\\
&GaitRef~\cite{zhu2023gaitref}  & {98.1} & {95.9} & {88.0} & 8.40M\\
\midrule
\multirow{4}{*}{Sil. + Joint + Bone} 
&GaitMix (w/ Bone) & 98.0 & 95.6 & 87.5 & -\\
&GaitRef (w/ Bone) & 98.2 & 96.0 & 88.9 & -\\
&\quad+ CMA$_{B\to J}$ & 98.1 & 95.7 & 88.7 & -\\
\cline{2-6}  \\ [-8pt]
&GaitSTR  & \textbf{98.4} & \textbf{96.2} & \textbf{89.6} & 11.48M\\
\bottomrule
\end{tabular}}
\end{center}
\end{table}

%% file: tables/abla-combination.tex
\begin{table}
\caption{Ablation results for different silhouette and skeleton feature combination on CASIA-B dataset for three splits. `Padding' indicates the skeleton feature is padded on each of the feature of different scales, while `concat.' means we concatenate the feature along with the scale dimension and use it only once.}
\label{table:abla-comb}
\begin{center}
{
\begin{tabular}{p{2.cm}p{2.cm}<{\centering}p{.8cm}<{\centering}p{.8cm}<{\centering}p{.8cm}<{\centering}}
\toprule
Method & Combination & NM & CL & BG \\
\midrule
GaitGL~\cite{lin2021gaitgl} & N/A & 97.3 & 94.4 & 83.5 \\
\midrule
Sil. + Joints & Padding & 97.5 & 94.6 & 85.8 \\
Sil. + Joints & Concat. & \textbf{98.1} & \textbf{95.9} & \textbf{88.0} \\
\bottomrule
\end{tabular}}
\end{center}
\end{table}

%% file: tables/abla-feat.tex
\begin{table}[t]
\caption{Ablations for different encoder and decoder combinations for silhouette with joints and different skeleton smoothing methods on CASIA-B datasets. Results are reported in Top-1 accuarcy.}
\begin{center}
{
\begin{tabular}{p{1.5cm}<{\centering}p{1.5cm}<{\centering}p{.8cm}<{\centering}p{.8cm}<{\centering}p{.8cm}<{\centering}}
\toprule
 Encoder & Decoder & NM & CL & BG \\
\midrule
 ST-GCN & N/A & 97.7 & 95.2 & 85.8  \\  %
 MS-G3D & N/A & 98.0 & 95.5 & 86.4 \\ %
\midrule   
 ST-GCN & ST-GCN & \textbf{98.1} & \textbf{95.9} & 88.0 \\ %
 ST-GCN & MS-G3D & \textbf{98.1} & 95.7 & \textbf{88.5} \\ %
 MS-G3D & ST-GCN & \textbf{98.1} & \textbf{95.9} & 88.3 \\ %
\midrule
 \multicolumn{2}{c}{Average Smoothing} & 97.6 & 95.0 & 85.6\\
 \multicolumn{2}{c}{Gaussian Smoothing} & 97.7 & 95.2 & 85.9 \\
 \multicolumn{2}{c}{SmoothNet \cite{zeng2021smoothnet}} & 97.4 & 94.4 & 83.8\\
\bottomrule
\end{tabular}}
\end{center}
\label{table:abla-feat}
\end{table}

\begin{table}[t]
\caption{Ablation results of different input for the skeleton correction network on CASIA-B. SCN is skeleton correction network.} 
\label{tab:pose}
\centering
\def\lw{1.4}
\def\lf{1.2}
\def\ls{0.05}
{
\begin{tabular}{p{\lw cm}<{\centering}p{\lf cm}<{\centering}p{\lw cm}<{\centering}p{\lw cm}<{\centering}} 
\toprule
Corr. Input & NM & BG &  CL  \\
\midrule
w/o $F_J$ & 97.7 & 95.4 & 87.0 \\
w/o $F_S$ & 97.6 & 95.3 & 85.6\\
w/o $J$ & 97.3 & 95.5 & 86.0\\
\midrule
Full SCN & \textbf{98.1} & \textbf{95.9} & \textbf{88.0}\\
\bottomrule
\end{tabular}
}
\end{table}

%% file: tables/ablasupp.tex
\begin{table}
\label{table:abla-supp}
\caption{Ablation results for $\lambda_1$, $\lambda_2$ and number of frames on CASIA-B.}
\begin{center}
\resizebox{\columnwidth}{!}
{\begin{tabular}{p{1.cm}<{\centering}p{1.cm}<{\centering}p{1.cm}<{\centering}p{.8cm}<{\centering}p{.8cm}<{\centering}p{.8cm}<{\centering}p{.8cm}<{\centering}}
\toprule
$\lambda_1$ & $\lambda_2$ & \# Frame & NM & CL & BG & Avg.\\
\midrule
1 & 1 & 30 & 98.4 & 96.2 & 89.6 & \textbf{94.7}\\
1 & 1 & 20 & 97.9 & 95.3 & 86.9 & 93.4\\
1 & 1 & 10 & 94.7 & 90.0 & 70.8 & 95.2\\
\midrule
0.01  & 1 & 30 & 96.7 & 92.2 & 76.5 & 88.5\\
0.1 &1 & 30 & 97.5 & 94.2 & 82.5 & 91.4\\
10 &1 & 30 & 98.0 & 95.8 & 90.1 & 94.6\\
\midrule
1&0.01 & 30 & 98.2 & 95.8 & 89.5 & 94.5\\
1&0.1  & 30 & 98.2 & 96.0 & 89.5 & 94.6\\
1&10   & 30 & 97.7 & 94.5 & 83.7 & 92.0\\
\bottomrule
\end{tabular}}
\end{center}
\end{table}

%% file: sections/5-conclusion.tex
\section{Conclusion}
We introduce \textit{GaitSTR}, building on GaitMix and GaitRef~\cite{zhu2023gaitref}, to integrate and refine skeletons with silhouettes for gait recognition. \textit{GaitSTR} incorporates bone representation alongside the joints, emphasizing the numerical connectivities between different nodes. It combines silhouettes and skeletons with two levels of refinement: silhouette-to-skeleton refinement for general guidance and dual-layer cross-modal adapters for sequential two-stream refinement between the joints and bones, ensuring temporal consistency across different representations. We compare \textit{GaitSTR} on four public datasets, including CASIA-B, OUMVLP, Gait3D, and GREW, and demonstrate state-of-the-art performance compare with other gait recognition methods.

%% file: sections/acknowledgement.tex
This research is based upon work supported in part by the Office of the Director of National Intelligence (ODNI), Intelligence Advanced Research Projects Activity (IARPA), via [2022-21102100007]. The views and conclusions contained herein are those of the authors and should not be interpreted as necessarily representing the official policies, either expressed or implied, of ODNI, IARPA, or the U.S. Government. The U.S. Government is authorized to reproduce and distribute reprints for governmental purposes notwithstanding any copyright annotation therein.